\documentclass{ws-ijmssc}
\usepackage{multirow}
\usepackage{subfigure}
\usepackage[graphicx]{realboxes}
\graphicspath{{images/}}

\begin{document}
\setlength{\parindent}{1em}

\markboth{Authors' Names}{Paper's Title}

\catchline{}{}{}{}{}

\title{A Multi-robot Coverage Path Planning Algorithm \\ Based on Improved DARP Algorithm }

\author{Yufan Huang\footnote{Sichuan University, No.24 South Section 1, Yihuan Road, Chengdu , China, 610065}}

\address{College of Electrical Engineering, Sichuan University\\Chengdu, China, 610065\\huangyufan1@stu.scu.edu.cn}

\author{Man Li}

\address{College of Electrical Engineering, Sichuan University\\Chengdu, China, 610065\\2020141440349@scu.edu.cn}

\author{Tao Zhao}

\address{College of Electrical Engineering, Sichuan University\\Chengdu, China, 610065\\
zhaotaozhaogang@126.com}

\maketitle

\begin{history}
\received{(Day Month Year)}
\accepted{(Day Month Year)}
\end{history}

\begin{abstract}
The research on multi-robot coverage path planning (CPP) has been attracting more and more attention. In order to achieve efficient coverage, this paper proposes an improved DARP coverage algorithm. The improved DARP algorithm based on A* algorithm is used to assign tasks to robots and then combined with STC algorithm based on Up-First algorithm to achieve full coverage of the task area. Compared with the initial DARP algorithm, this algorithm has higher efficiency and higher coverage rate.
\end{abstract}

\section{Introduction}	
Robot full coverage is a path planning problem for one or more mobile robots to traverse the target environment using sensor information. It needs to optimize goals such as high coverage rate, low coverage time, low repeat coverage rate, and low energy consumption \cite{1}. Robot full coverage technology is widely used in military, agriculture, industry, rescue and life fields. At present, research in the field of single robot full coverage is relatively mature. This technology has been put into use in low-end and medium-end robot industries such as household sweeping robots and industrial inspection robots. Due to the complexity of task scenarios and the increase in area size, single robot full coverage often cannot meet the requirements. Thus Multiple robots are needed to work together to efficiently and conveniently complete the coverage task.

The problem of multi-robot coverage involves environmental area division, task assignment and path planning algorithms. At present, the methods for dividing the task space mainly include grid decomposition method \cite{2}, topological logic diagram\cite{3}, Morse diagram \cite{4}, Voronoi diagram \cite{5}, etc. The main path planning algorithms for multi-robot full coverage include tree-based path planning \cite{6}, greedy search and graph search algorithms \cite{7}, genetic algorithms \cite{8}, etc. A core problem of multi-robot coverage is task assignment. Task assignment algorithms are mainly divided into heuristic assignment methods \cite{9}, market auction-based assignment methods \cite{10}, and swarm intelligence-based methods \cite{11}, etc. These related algorithms usually only consider major cover-age issues such as high coverage rate and low repeat rate, which makes it difficult to achieve uniform task distribution and good robustness.

In 2017, Athanasios et al. \cite{12} proposed a DARP (Divide Areas based on Robots Initial Positions) algorithm. It divides the task area based on the initial positions of robots, and plans the coverage path with STC(spanning tree coverage) algorithm to find the optimal solution that satisfies constraints such as coverage rate, non-backtracking, and balanced distribution.

However, there are still some problems with this algorithm. First of all, the task area division of the DARP algorithm has limitations, which increases the number of turns of the robot and thus increases the coverage time. Secondly, since the premise of the generation tree algorithm assumes that the obstacle area division is based on four times the size of the robot as the basic unit, this makes the basic unit of obstacle area division too large, resulting in a larger area of obstacle area division than the actual area of obstacle area, which creates a “blind spot” and cannot achieve complete coverage.

Based on these problems, we use A* algorithm to optimize DARP algorithm evaluation matrix so that task assignment is more reasonable and uniform. Moreover, we propose UF(Up-First) algorithm to make up for the limitation of STC algorithm’s inability to cover small cells by changing the basic unit of obstacle area from four times the robot size to one times the robot size. Through the simulation analysis of this algorithm in this article, it can be concluded that this algorithm can effectively reduce the number of turns of the robot, shorten the coverage time, reduce energy consumption, and improve the coverage rate, truly achieving complete coverage.

In summary, the innovations proposed in this paper mainly include:
\begin{romanlist}[(ii)]
\item using the A* algorithm to improve the DARP algorithm to optimize the evaluation matrix to make the task allocation more reasonable; 
\item proposing the UF-STC algorithm to optimize the cover-age path, so that the robot can achieve 100\% coverage with higher accuracy.
\end{romanlist}
The details of the multi-robot CPP researches are separated in several sections, and the outline of this paper is as fol-lows. Sec.2 mainly focuses on technical details of A*-DARP algorithm with Sec.2.1 about prepared knowledge, Sec.2.2 about the DARP algorithm, and Sec.2.3 about the improve-ment based on A* algorithm. Sec.3 presents improvement of STC algorithm with Sec.3.1 about STC algorithm, and Sec.3.2 about the UF algorithm.  Sec.4 presents the research data and related analyses. Finally in Sec.5 a conclusion and future re-search topics are mentioned.
\section{Improved DARP Algorithm}
DARP algorithm is a method for assigning coverage tasks to robots based on their initial positions. This paper improves it by using A* algorithm, which can effectively reduce the number of turns in path planning, increase coverage efficiency, and decrease loss.

\subsection{Prepared knowledge}
\subsubsection{Grid map model}
The representation method of environmental map is map model. A good map model should meet the following requirements: 
\begin{romanlist}[(ii)]
\item easy for computer processing; 
\item easy to update maps with new information; 
\item mobile robots can rely on the map to complete specific tasks, such as positioning, path planning, etc.
\end{romanlist}
We adopt grid method to build map model, which was first proposed by Elfes and Moraves \cite{2} and has been applied in many robot systems.

The specific content of the grid method is to determine the size of each grid in the grid map according to the shape parameters of the robot, and then divide the environment map according to the size to obtain the grid map. The size of the grid will directly affect the memory required by the computer to store the environment information, and will also directly affect the time required for calculation when planning the path. Therefore, the grid size should not be too large or too small.

The grid model is easy to create and maintain. The information of each cell that the robot knows directly corresponds to a certain region in the environment. With the help of the grid map, the robot can easily carry out self-localization, path planning and other intelligent decision-making behaviors \cite{13}.

In this paper, we decompose the real environment map into a grid map whose grid length is the length of a robot and width is the width of a robot. In particular, in the task allocation process, the environment map needs to be de-composed into a grid map with the length of 2*length and the width of 2*width, that is, four adjacent grids are merged into a large grid in the initial grid map.

\subsubsection{A* algorithm}
A* algorithm \cite{14} \cite{15} is a heuristic search algorithm, which combines the advantages of Dijkstra algorithm \cite{17} and best-first search algorithm.

The function expression of A* algorithm is
\begin{equation}
f(n) = g(n) + h(n)
\end{equation}
where \(g(n)\) denotes the cost distance from the starting node to the current node n, \(h(n)\) denotes the estimated cost distance from the current node n to the ending node \cite{14}\cite{15}, and \(f(n)\) is the integrated cost distance of node n. In order to find the shortest path from the start node to the end node, the node which has the smallest \(f(n)\) is selected each time. So the deciding factor is \(f(n)\). Since the cost distance from the starting node to the current node is determined, it is very important to choose the appropriate estimated cost distance \(h(n)\) for the whole search process \cite{18}. The steps of the A* algorithm are described in Table 1.

\begin{table}[ht]
\tbl{ A* algorithm}
{\begin{tabular}{ll}
\hline
\textbf{Input}  & \textbf{map information, robot’s start point and end point}                                                                                             \\
\textbf{Output} & \textbf{the shortest path from start point to end point}                                                                                                \\
\textbf{Step 1} & \begin{tabular}[c]{@{}l@{}}Initialize open set and closed set, add start point to open \\ set\end{tabular}                                              \\
\textbf{Step 2} & Select the node n with the smallest total cost distance from open set                                        \\
\textbf{Step 3} & If n is not the end point, update open set and closed set                                                  \\
\textbf{Step 4} & \begin{tabular}[c]{@{}l@{}}Traverse the adjacent nodes of n, add those that are not in closed set \\ to open set, and set n as parent node\end{tabular} \\
\textbf{Step 5} & Repeat step 2 to step 4 until n is the end point                                                                                                        \\
\textbf{Step 6} & \begin{tabular}[c]{@{}l@{}}If n is the end point, trace back parent nodes until start point and \\ output path\end{tabular}                             \\ \hline
\end{tabular}}
\label{table1}
\end{table}

\subsection{DARP algorithm for task allocation}
DARP algorithm \cite{12} divides the environment map into several independent single-robot task areas and then simplifies the multi-robot CPP task to several single-robot path planning tasks. This greatly reduces the difficulty of the multi-robot CPP task.

First, for each robot i, an evaluation matrix \(E_i\) is generated. The evaluation matrix represents the reachability of the initial position of the i-th robot to other cells in \(L\) on the map, where L is the set of cells that need to be covered on the map. Then an assignment matrix \(A\) is introduced to determine the ownership relationship of each cell in the task allocation process:
\begin{equation}
A_{x, y}=\underset{i \in\left\{1, \ldots, n_r\right\}}{\arg \min } E_{i \mid x, y}, \forall x, y \in L
\end{equation}
The task area assigned to each robot can be expressed as
\begin{equation}
L_i=\{(x, y) \in L: A(x, y)=i\}, \forall i \in\left\{1, \ldots, n_r\right\}
\end{equation}
The number of cells assigned to the i-th robot can be expressed as
\begin{equation}
k_i=\left|L_i\right|, \forall i \in\left\{1, \ldots, n_r\right\}
\end{equation}
Before the iterative process begins, the evaluation matrix only contains information about the initial cost from the robot position to other cells, that is
\begin{equation}
E_{i \mid x, y}=d\left(\chi_i\left(t_0\right),[x, y]^\tau\right), \forall i \in\left\{1, \ldots, n_r\right\}
\end{equation}
where \(d()\) represents the Euclidean distance from the robot position to other cells.

In order to ensure that each robot is assigned to a roughly equal task area, the algorithm introduces an evaluation function \(J\) to describe the fairness of task allocation:
\begin{equation}
J=\frac{1}{2} \sum_{r=1}^{n_r}\left(k_i-f\right)^2
\end{equation}
It is clear that the task allocation plan is most equitable when the cells to be covered are equally assigned to each robot, thus
\begin{equation}
f=L/n_r
\end{equation}
In order to optimize \(E_i\) continuously during the iterative process, DARP algorithm introduces a correction factor \(m_i\), which makes
\begin{equation}
E_{i}=m_{i}E_{i}
\end{equation}
\(m_i\) is called the scalar correction factor for the i-th robot.

To achieve the minimum of the evaluation function \(J\), coordinate gradient descent [19] can be applied to optimize the scalar correction factor: 
\begin{equation}
m_i=m_i-\eta \frac{\partial J}{\partial m_i}
\end{equation}
From this we can obtain the optimal solution vector \(m^*\), such that
\begin{equation}
J\left(m^*\right) \leq J(m), \forall m \in \operatorname{dom}(J)
\end{equation}
We can then obtain the optimal evaluation matrix \(E_{i}\).

Although the above process can make the task area as-signed to each robot optimal, it cannot guarantee that the task area is connected. To solve this problem, we introduce the following matrix 
\begin{equation}
\begin{gathered}
C_{i \mid x, y}=\min (\|[x, y]-r\|)-\min (\|[x, y]-q\|) \\
\forall r \in R_i, q \in Q_i
\end{gathered}
\end{equation}
where \(R_i\) denotes the region where the initial position \(\chi_i\left(t_0\right)\)  of the i-th robot is located, and \(Q_i\) denotes the region assigned to the i-th robot but not connected to \(\chi_i\left(t_0\right)\). It can be seen that the role of  \(C_i\) is to reward the regions connected to \(\chi_i\left(t_0\right)\) and penalize the regions not connected to \(\chi_i\left(t_0\right)\) , thus achieving the result that all task regions of the i-th robot are connected. The evaluation matrix after introducing \(C_i\) becomes
\begin{equation}
E_i=C_i \odot\left(m_i E_i\right)
\end{equation}
where \(\odot\) denotes element-wise multiplication. At this time, \(E_i\) satisfies all the requirements of multi-robot task allocation.

\subsection{Improvement based on A* algorithm}
DARP algorithm introduced above can effectively complete the task allocation of multiple robots. The evaluation matrix obtained by calculating the Euclidean distance can perform well in generating the initial evaluation matrix in most cases, such as when the obstacles are not dense, or when the area near the initial positions of the robots is relatively open, that is, there are no obstacles nearby, or when the obstacles are not connected in a line.

Considering the physical limitations of robots, turning during coverage will consume more time than straight-line motion, resulting in more energy and mechanical losses. Therefore, reducing the number of turns of robots can effectively improve their coverage efficiency.

\begin{figure}[ht]
    \centering
    \includegraphics[width=5cm,height=5cm]{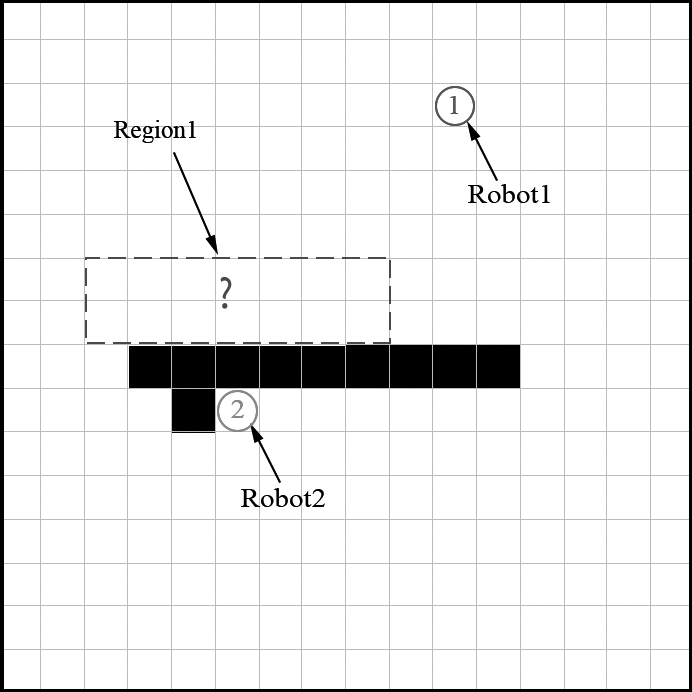}
    \caption{Difficult scenario for DARP algorithm}
    \label{Fig1}
\end{figure}

Figure 1 shows a scenario that DARP algorithm will en-counter. In figure 1, Robot 1 and Robot 2 represent the initial positions of robots, Region 1 represents a difficult region for DARP to handle.

When the situation in figure 1 occurs, \(E_i\) generated by calculating Euclidean distance will assign Region 1 to Robot 2. However, due to obstacle blocking, it is more reasonable to assign Region 1 to Robot 1 because Robot 1 can reach that region more easily than Robot 2.

After Region 1 is assigned to Robot 2, more turns will be generated during spanning tree construction and path formation process which greatly increases coverage time and produces more losses.

To fix this problem, this paper proposes an improved DARP algorithm based on A* algorithm. Specifically, it uses A* algorithm’s advantages in obstacle avoidance and finding shortest path to optimize evaluation matrix generation and achieve more reasonable task allocation. The improvement process is as follows.

We use A* algorithm to calculate the shortest path from the i-th robot to all other reachable cells in the map. The robots are constrained to move orthogonally, so we choose the Manhattan distance from the initial point to the target end-point as the estimated cost distance \(h(n)\) for A* algorithm, that is 
\begin{equation}
h(n)=\left|x_1-x_2\right|+\left|y_1-y_2\right|
\end{equation}
The set of shortest paths obtained forms the initial evaluation matrix for the i-th robot, that is
\begin{equation}
E_{i \mid x, y}=A\left(\chi_i\left(t_0\right),[x, y]^\tau\right), \forall i \in\left\{1, \ldots, n_r\right\}
\end{equation}
where \(A()\) denotes the path distance from the robot position to other cells calculated by A* algorithm.

\begin{figure}[htb]
    \centering
    \subfigure[path of DARP algorithm]{
    \label{Fig.sub.1}
    \includegraphics[width=5cm,height=5cm]{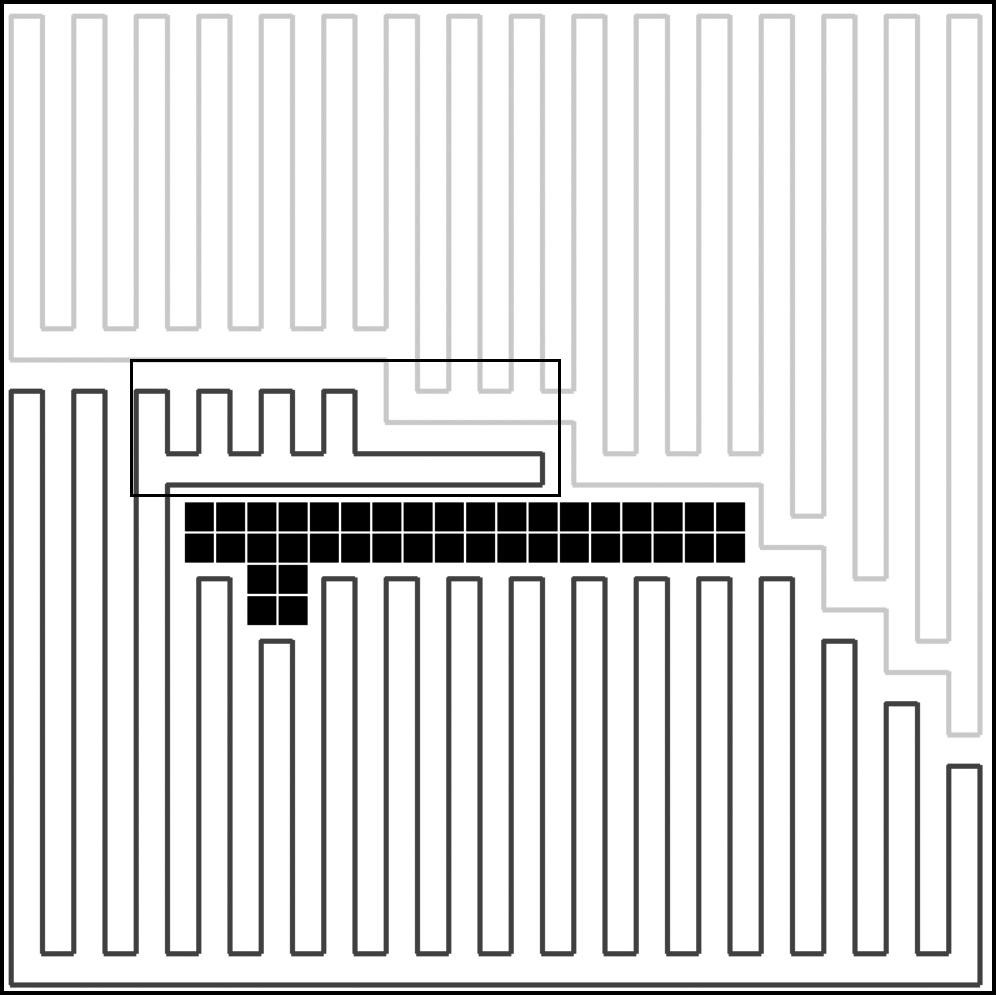}}
    \subfigure[path of A*-DARP algorithm]{
    \label{Fig.sub.2}
    \includegraphics[width=5cm,height=5cm]{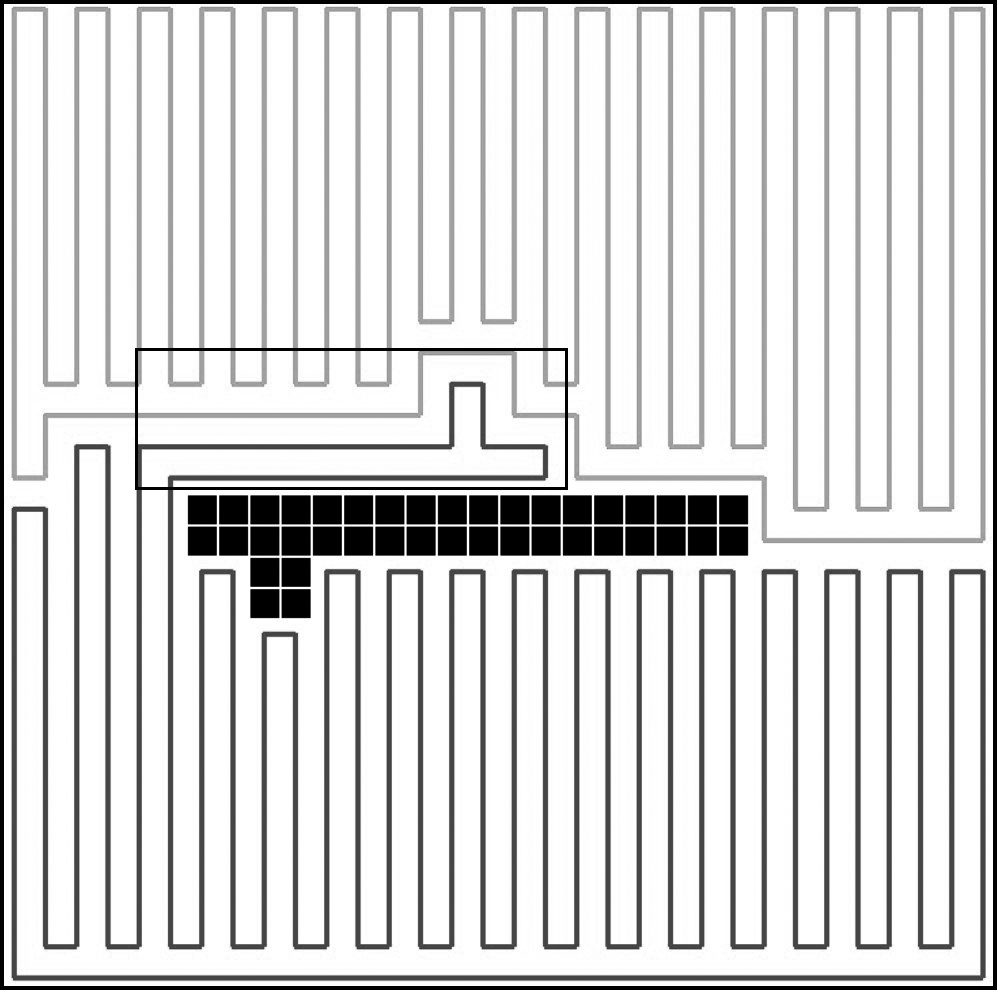}}
    \caption{comparison of the two algorithm}
    \label{Fig2}
\end{figure}

Figure 2 compares the paths formed by using Euclidean distance as the evaluation matrix and using A* algorithm to improve the evaluation matrix. From Figure 2a and 2b, it can be seen that the improved DARP algorithm effectively reduces the number of turns of the robots in the highlighted area, greatly saving the robot’s coverage time, as well as the energy and mechanical loss caused by turning.

\section{UF Algorithm for CPP of a Single Robot}
After obtaining the task area of each robot, we transform the task into   single robot coverage problem. We use the minimum spanning tree algorithm based on grid method to solve this problem, and use UF algorithm to make up for the short-comings of the minimum spanning tree algorithm.

\subsection{STC algorithm}
STC algorithm \cite{20} \cite{21}  can construct a covering all the task areas, starting from any unoccupied cell, and it can achieve 100\% path coverage under the condition of grid map.

After a single robot obtains the assigned task area divided by the grid method, the Kruskal algorithm \cite{22}  is used to generate the minimum spanning tree. Robots move along the minimum spanning tree counterclockwise or clockwise to achieve complete coverage of a single robot in the task area.

\begin{figure}[htb]
    \centering
    \includegraphics[width=5cm,height=5cm]{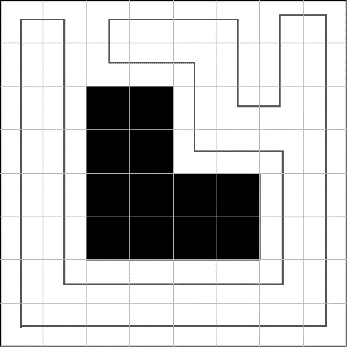}
    \caption{Example of a path generated by STC}
    \label{Fig3}
\end{figure}

Figure 3 illustrates the spanning tree path for a single robot, where it can be seen that every cell marked as a feasible region is covered, achieving 100\% coverage.

\subsection{UF algorithm}
Although STC algorithm is able to achieve 100\% coverage path planning under grid map conditions, due to the limitation of grid map, it cannot achieve 100\% coverage in the perspective of real map. Specifically, the grid map method merges four small cells into one large cell for DARP task allocation and spanning tree construction. When one of the four small cells is an obstacle in the real map, the large cell will be marked as an obstacle regardless of the feasible states of the other three small cells in the real map. As a result, small cells which are marked as obstacles in the grid map, but are feasible areas in the real map cannot be covered by the path planned by the spanning tree, and the coverage rate is reduced. For convenience of presentation, these cells are called ‘small cells’.

Therefore, this paper proposes an Upward-First path compensation algorithm to overcome the shortcomings of the spanning tree algorithm based on grid method. The following are the basic principles of this algorithm.

Assume that the robot has four motion states in the process of forming a path around the spanning tree: up, left, down, and right (the directions here are absolute directions based on the map, not relative directions of the robot during motion). We assign corresponding priorities to these motion states. This paper adopts an Upward-First method, with priorities in order: up, left, down, right.

During the process of constructing the path along the minimum spanning tree, if the robot detects the existence of a small cell around the current cell, the robot will branch out of the spanning tree path and move one cell towards that cell, while continuing to detect the existence of small cells accord-ing to the priority list. If there is one, it will continue to move towards the small cell. The robot will continue to move until it reaches a dead zone along the current direction, where all four directions of the current cell are blocked (obstacle blocking, boundary blocking, covered area blocking, etc.).

Then, it will return in the opposite direction of the current direction, and if it detects a small cell during the return, it will branch out in the same way and move towards that cell for coverage. The robot will continue to move until it returns to the spanning tree path and continues to cover along the minimum spanning tree.

\begin{figure}[htb]
    \centering
    \includegraphics[width=5cm,height=5cm]{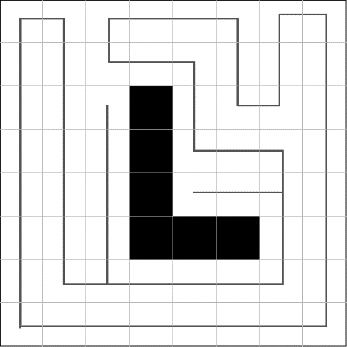}
    \caption{Example of a path generated by UF-STC}
    \label{Fig4}
\end{figure}

Figure 4 shows the coverage path of STC algorithm improved by UF algorithm on a map with small cells. From the above description and pictures, it can be seen that STC algorithm cannot handle the coverage problem of small cells, while UF algorithm can cover the small cell area 100\% with-out affecting the path formed by the spanning tree. This can greatly improve the coverage rate and coverage accuracy. The steps of UF algorithm are shown in Table 2.
\begin{table}[ht]
\tbl{ UF algorithm}
{\begin{tabular}{ll}
\hline
\textbf{Input}    & \textbf{Real   map information, robot position parameters}                                                                                                                                                                       \\
\textbf{Output}   & \textbf{Optimized   coverage path}                                                                                                                                                                                               \\
\textbf{Step   1} & \begin{tabular}[c]{@{}l@{}}Detect   the existence of small cells according to the priority order \\ of up, left, down, and right. If there is one, move in that direction.\\  If all directions are blocked, return\end{tabular} \\
\textbf{Step   2} & \begin{tabular}[c]{@{}l@{}}Continue   to detect small cells according to the priority order \\ during the return. If there is one, branch out from the current \\ path and move towards that direction\end{tabular}              \\
\textbf{Step   3} & \begin{tabular}[c]{@{}l@{}}Return   to the spanning tree path and continue to construct the \\ path along the spanning tree\end{tabular}                                                                                         \\ \hline
\end{tabular}}
\end{table}

\section{Result Analysis}
This paper compares the proposed algorithms, obtains simulation results, and analyzes the advantages and disadvantages of the algorithm. The evaluation of algorithm performance is divided into two groups. One group is a comparison between A*-DARP algorithm and DARP algorithm. The other group is a comparison between UF-STC algorithm and STC algorithm.

\subsection{The evaluation of A*-DARP algorithm performance}

This paper simulates multiple different experimental map models, and a large number of simulation results show that the algorithm proposed in this paper has better performance. In order to better demonstrate the superiority of this paper’s algorithm, we selected three maps with different obstacle ratios, planned full coverage paths for different obstacle map environments with 3-5 robots, and analyzed the operation data.
\begin{figure}[htb]
    \centering
    \subfigure[]{
    \label{Fig5.sub.1}
    \includegraphics[width=3cm,height=3cm]{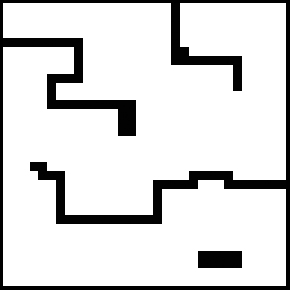}}
    \subfigure[]{
    \label{Fig5.sub.2}
    \includegraphics[width=3cm,height=3cm]{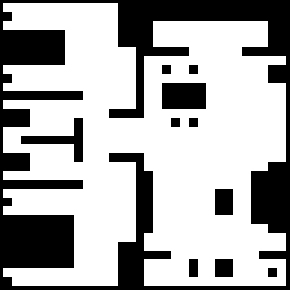}}
     \subfigure[]{
    \label{Fig5.sub.3}
    \includegraphics[width=3cm,height=3cm]{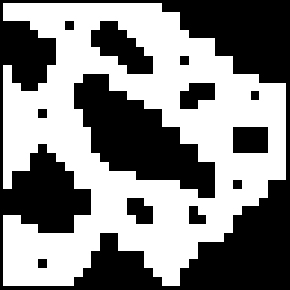}}
    \caption{gird map used for simulation}
    \label{Fig5}
\end{figure}

As shown in Figure 5, the map environment consists of a 64*64 grid map, and the effective coverage range of a single robot is equivalent to the size of one grid. The white grid parts shown in the map are obstacle-free areas, that is, feasible are-as, and the black parts are obstacle areas, that is, non-feasible areas. The three maps have different obstacle ratios to test the performance of the algorithm under different map conditions. The obstacle ratio of Map a is 10.1\%, Map b is 29.6\%, and Map c is 43.5\%. Each map is covered by a cluster of robots with initial positions uniformly distributed.

Four evaluation indicators are used in the experiment. Max means time spent by the robot that takes longest to complete the coverage task; Min means time spent by the robot that takes shortest time to complete the coverage task; Ave means average time spent by each robot to complete the coverage task; Ratio means ratio of maximum coverage time to minimum coverage time. Where the time taken by the first robot to complete the coverage task
\begin{equation}
T_{ci}=n_it_s+m_it_t, i \in\left\{1, \ldots, n_r\right\}
\end{equation}
where \(t_s\) represents time taken by a robot to move in a straight line between two adjacent grids; \(t_t\) represents time taken by a robot to turn 90° and move forward one grid; \(n_i\) represents number of straight-line movements required for a robot to complete coverage task; \(m_i\) represents number of 90° turns required for a robot to complete coverage task. Ac-cording to actual measurements, \(t_s\) is considered as one unit time and \(t_t\) as 1.5 units of time. The simulation results of both algorithms are shown in Table 3. Figure 6 shows the coverage performance of both algorithms on different test maps.

\begin{table}[ht]
\tbl{ analysis of simulation data}
{\begin{tabular}{cccccccccc}
\hline
\multirow{2}{*}{\textbf{map}} & \multirow{2}{*}{\textbf{robots}} & \multicolumn{4}{c}{DARP+STC} & \multicolumn{4}{c}{A*-DARP+STC} \\
                              &                                  & Max  & Min  & Ave    & Ratio & Max   & Min  & Ave     & Ratio  \\ \hline
\textbf{a}                    & 3                                & 175  & 135  & 155.00 & 1.30  & 155   & 143  & 143.67  & 1.08   \\
\textbf{}                     & 4                                & 136  & 68   & 112.25 & 2.00  & 136   & 68   & 107.25  & 2.00   \\
\textbf{}                     & 5                                & 131  & 59   & 93.40  & 2.22  & 109   & 59   & 86.60   & 1.85   \\
b                             & 3                                & 185  & 141  & 161.67 & 1.31  & 169   & 141  & 157.67  & 1.19   \\
                              & 4                                & 145  & 95   & 118.25 & 1.52  & 129   & 89   & 111.25  & 1.45   \\
                              & 5                                & 141  & 101  & 122.00 & 1.40  & 133   & 103  & 118.00  & 1.29   \\
c                             & 3                                & 171  & 146  & 155.00 & 1.17  & 167   & 144  & 153.00  & 1.16   \\
                              & 4                                & 147  & 113  & 124.50 & 1.30  & 139   & 109  & 123.50  & 1.28   \\
                              & 5                                & 135  & 98   & 109.60 & 1.38  & 127   & 96   & 106.40  & 1.32   \\ \hline
\end{tabular}}
\end{table}

\begin{figure}[htb]
    \centering
    \subfigure[DARP algorithm simulation result 1]{
    \label{Fig6.sub.1}
    \includegraphics[width=5cm,height=5cm]{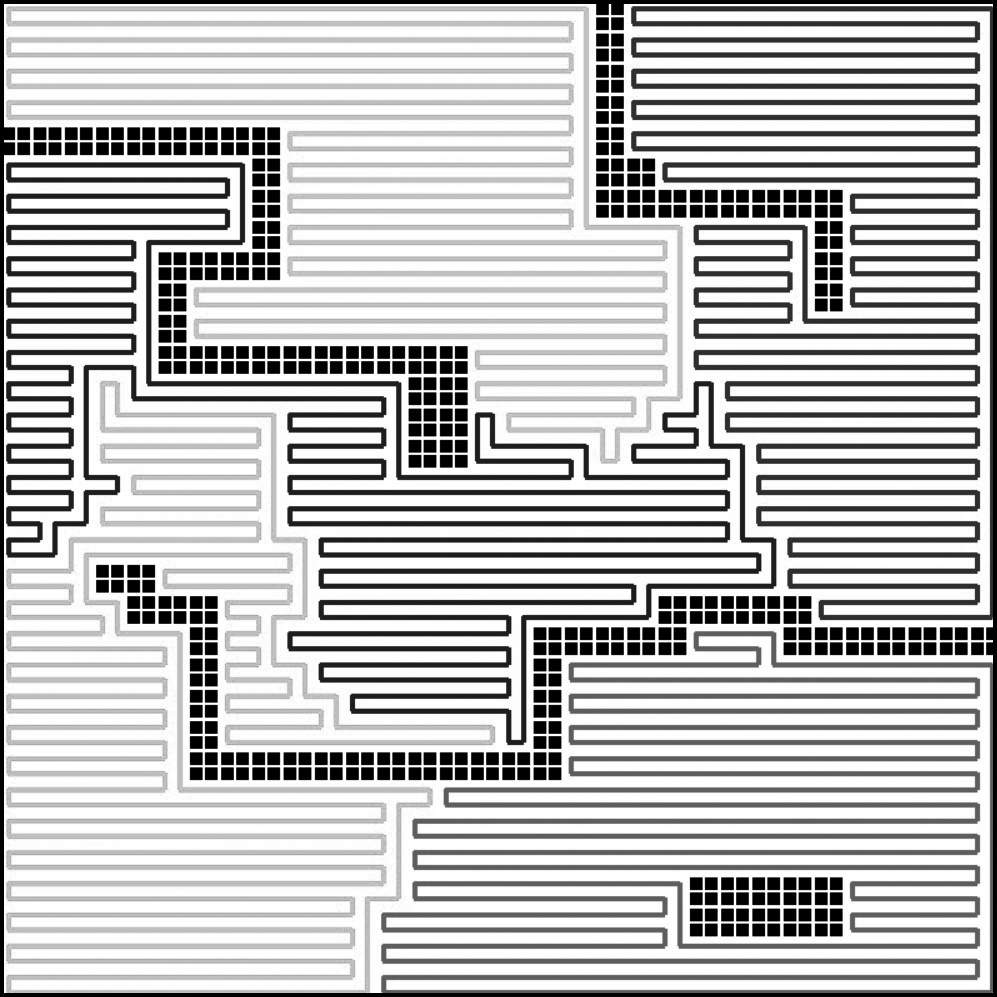}}
    \subfigure[A*-DARP algorithm simulation result 1]{
    \label{Fig6.sub.2}
    \includegraphics[width=5cm,height=5cm]{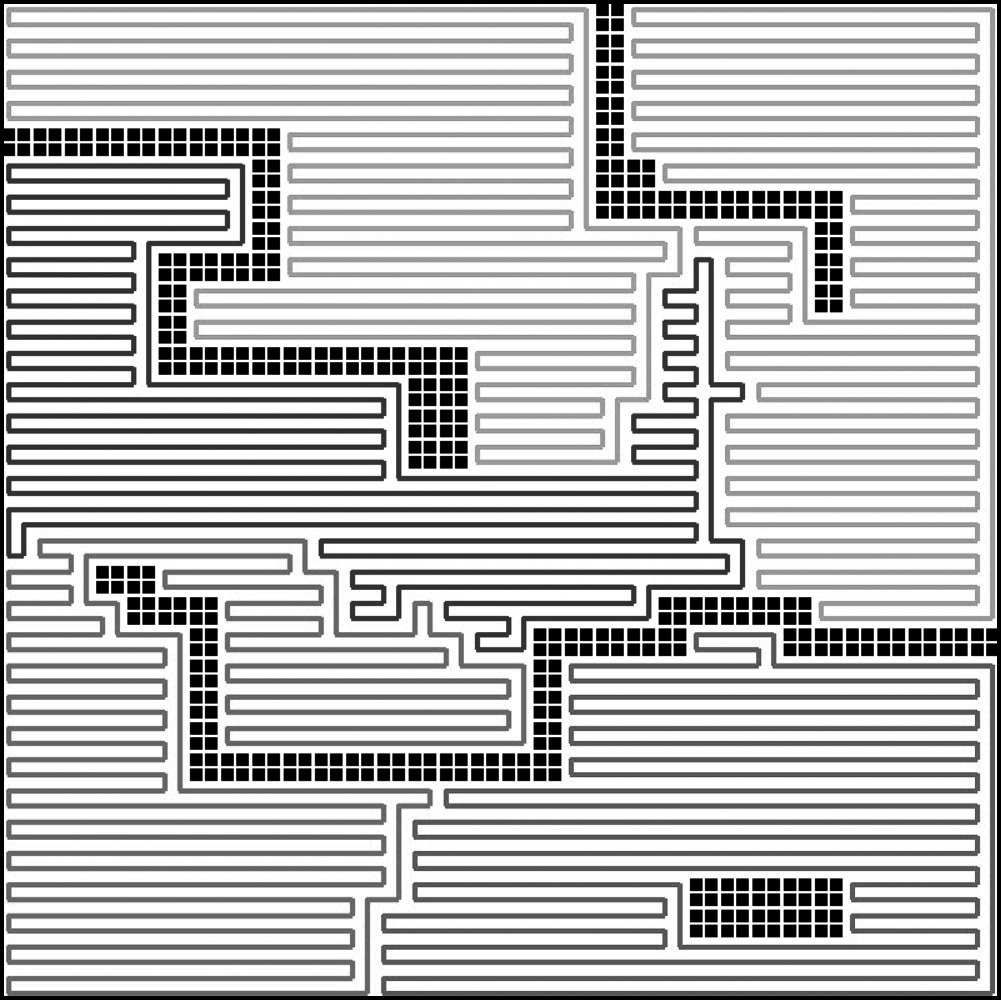}}
     \subfigure[DARP algorithm simulation result 2]{
    \label{Fig6.sub.3}
    \includegraphics[width=5cm,height=5cm]{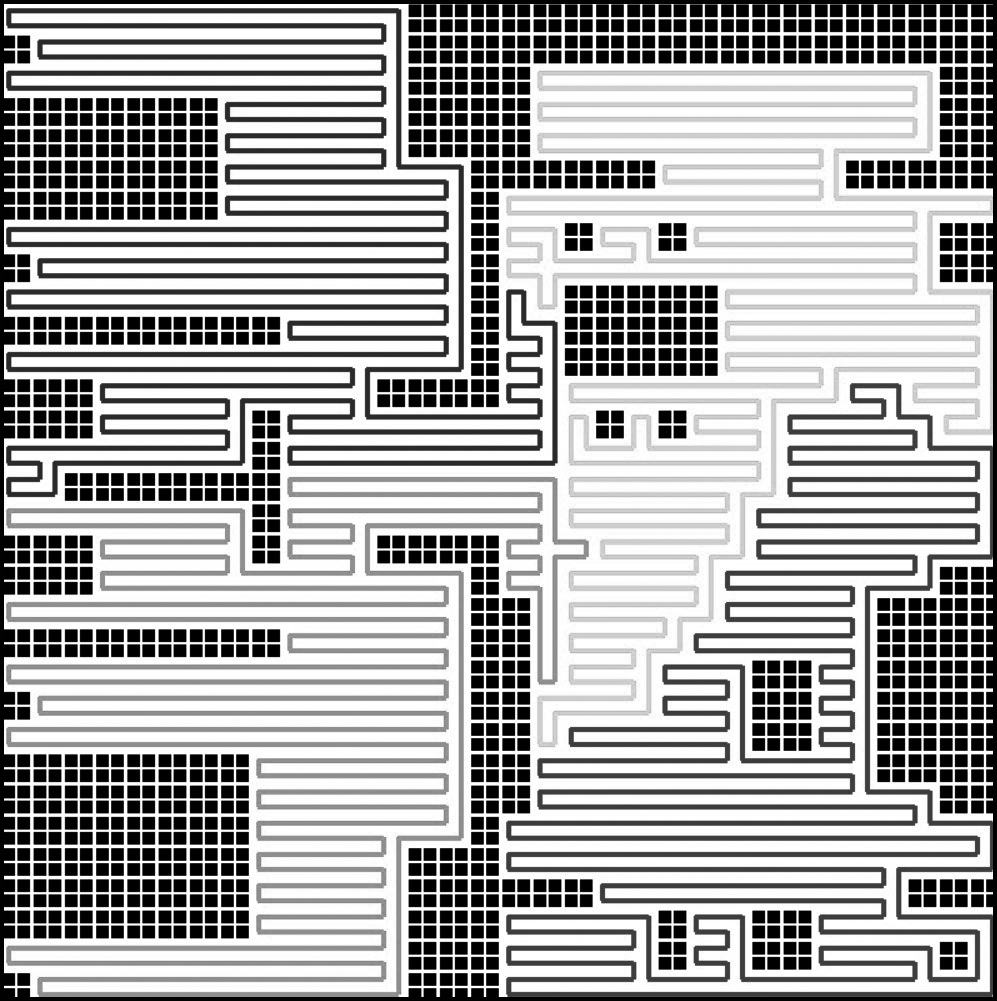}}
     \subfigure[A*-DARP algorithm simulation result 2]{
    \label{Fig6.sub.4}
    \includegraphics[width=5cm,height=5cm]{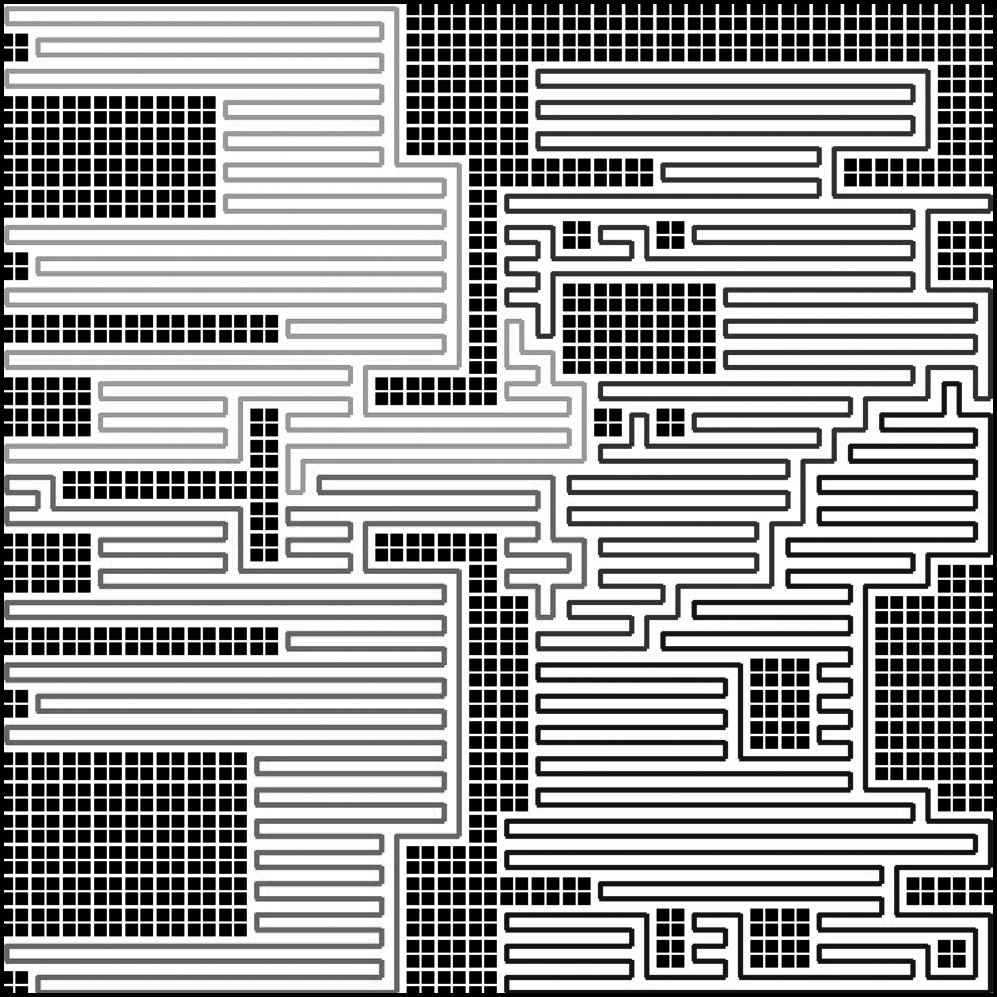}}
     \subfigure[DARP algorithm simulation result 3]{
    \label{Fig6.sub.5}
    \includegraphics[width=5cm,height=5cm]{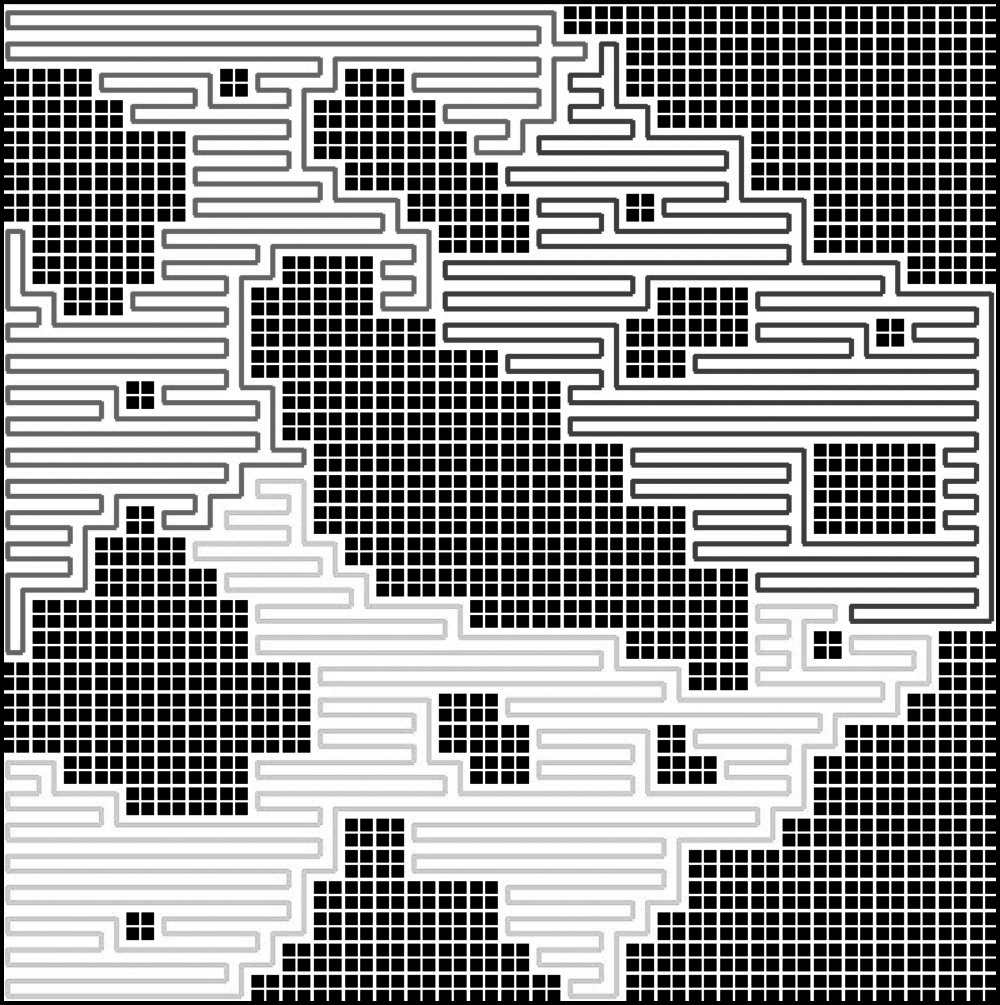}}
     \subfigure[A*-DARP algorithm simulation result 3]{
    \label{Fig6.sub.6}
    \includegraphics[width=5cm,height=5cm]{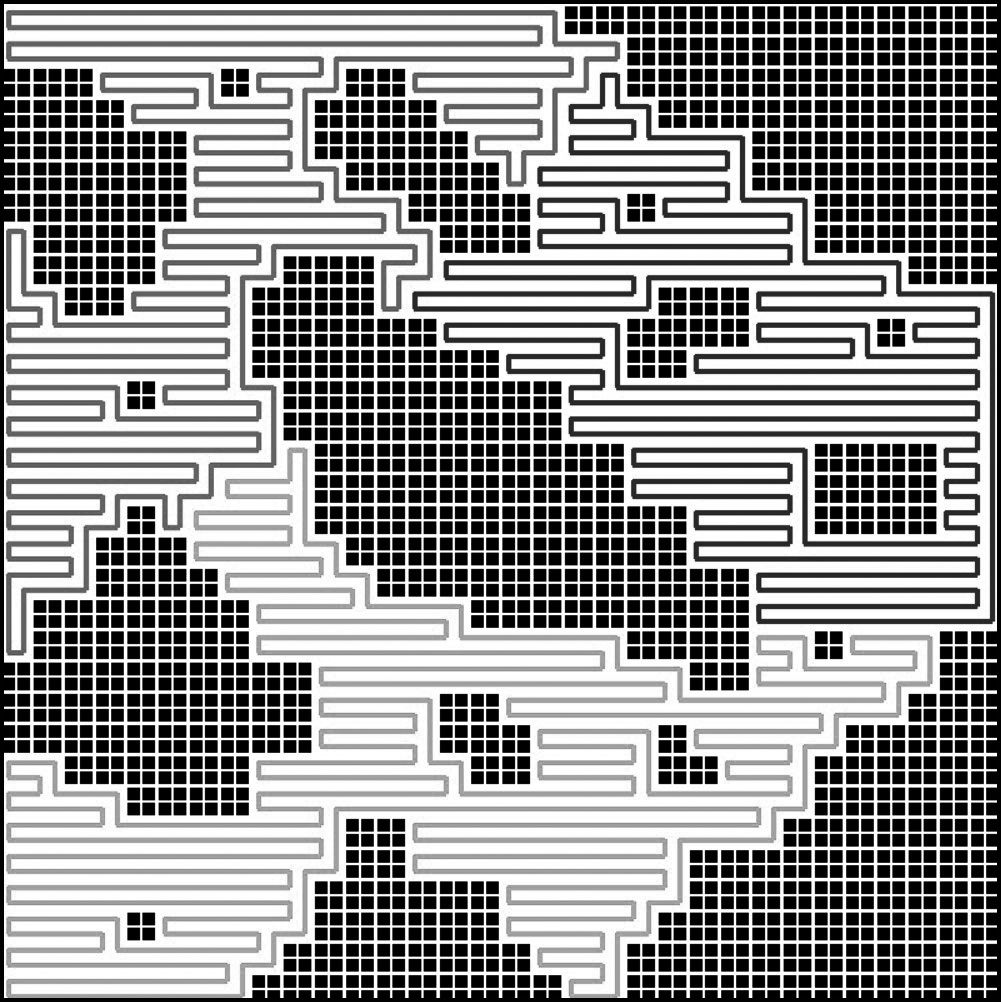}}
    \caption{DARP algorithm/A*-DARP algorithm simulation results}
    \label{Fig6}
\end{figure}
Table 3 and Figure 6 show that under various types of map conditions, compared with DARP algorithm, A*-DARP algorithm has fewer turns for robots which can greatly improve coverage efficiency and reduce losses, achieving full coverage goals in shorter time. Coverage tasks assigned to individual robots are more evenly distributed and each robot can complete its own coverage task in similar times which can further make full use of high efficiency advantage of multi-robot work and improve efficiency of multi-robot collaboration.

\subsection{The evaluation of UF-STC algorithm performance}

In order to verify the effectiveness of the UF-STC algorithm, we select maps with small cells as described earlier for testing. The map environment consists of a 32*32 grid map, and the obstacle ratio of Map d is 13.8\%, and that of Map e is 22.3\%.

\begin{figure}[htb]
    \centering
    \subfigure[STC algorithm simulation result 1]{
    \label{Fig7.sub.1}
    \includegraphics[width=5cm,height=5cm]{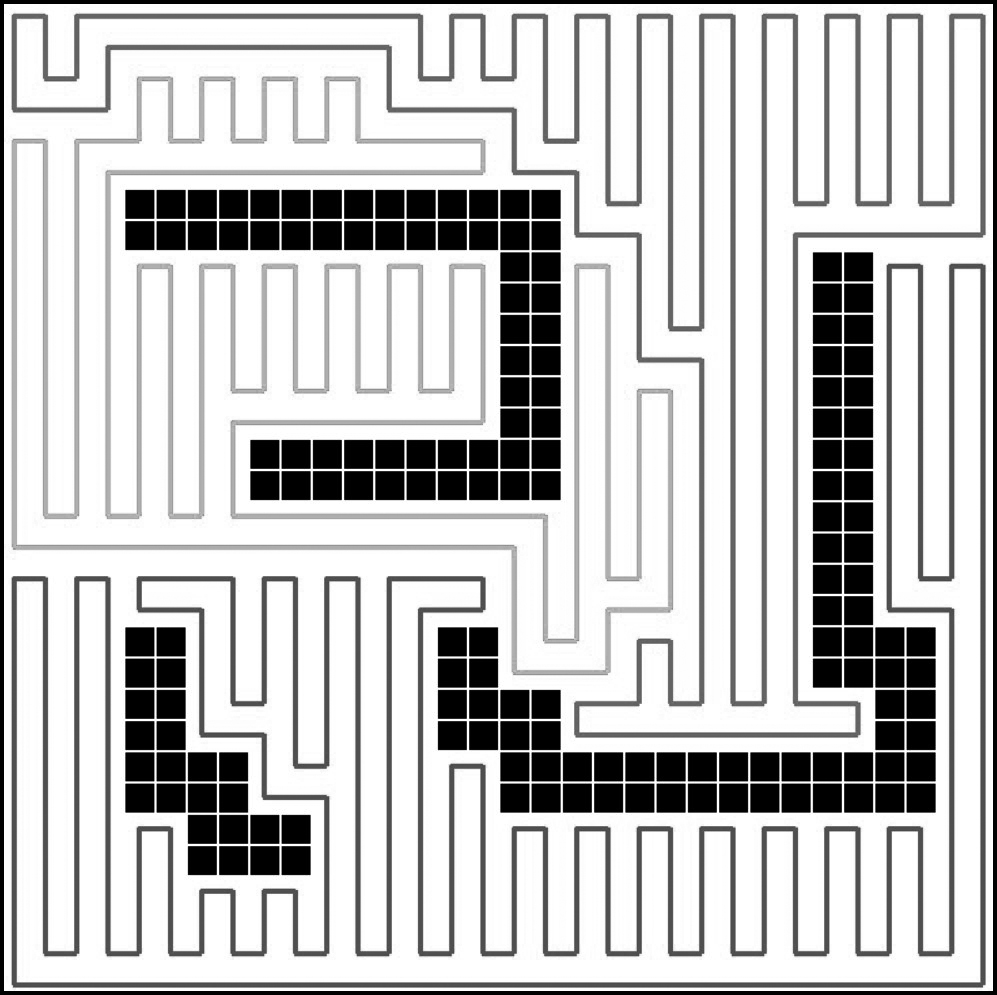}}
    \subfigure[UF-STC algorithm simulation result 1]{
    \label{Fig7.sub.2}
    \includegraphics[width=5cm,height=5cm]{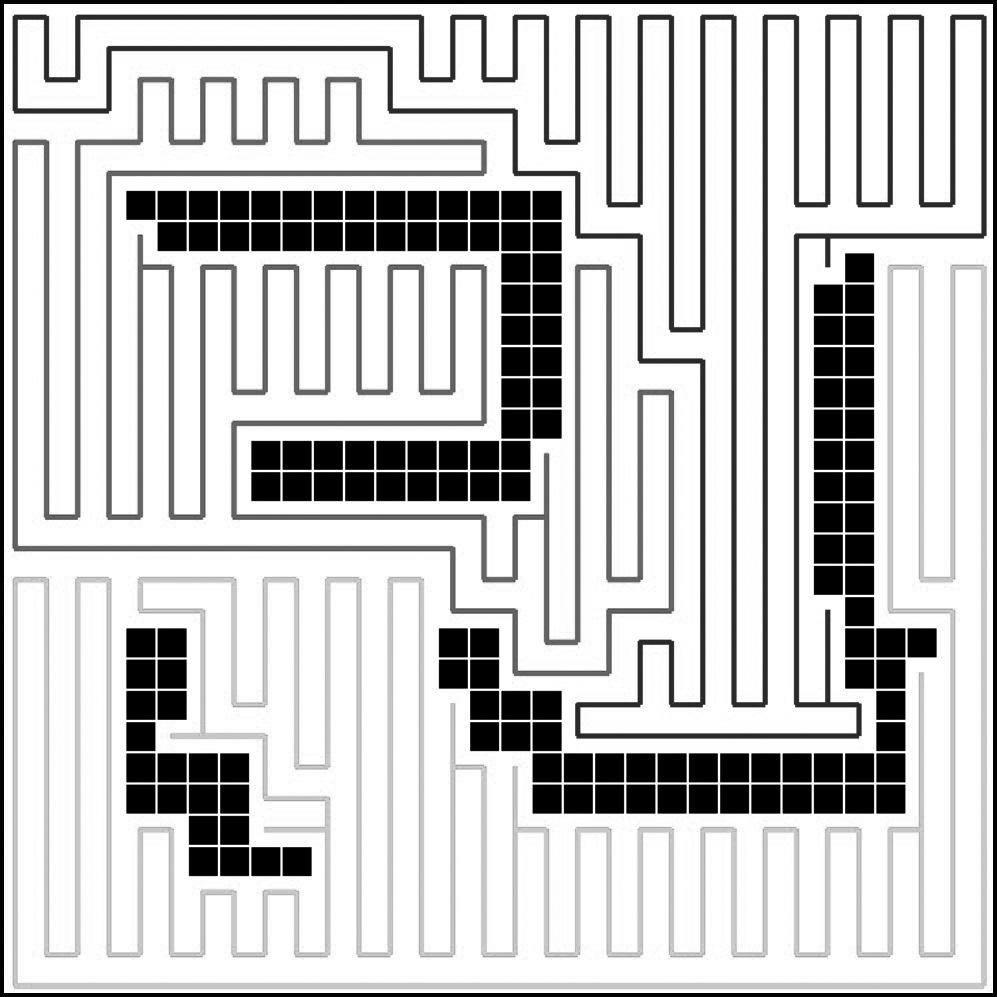}}
     \subfigure[STC algorithm simulation result 2]{
    \label{Fig7.sub.3}
    \includegraphics[width=5cm,height=5cm]{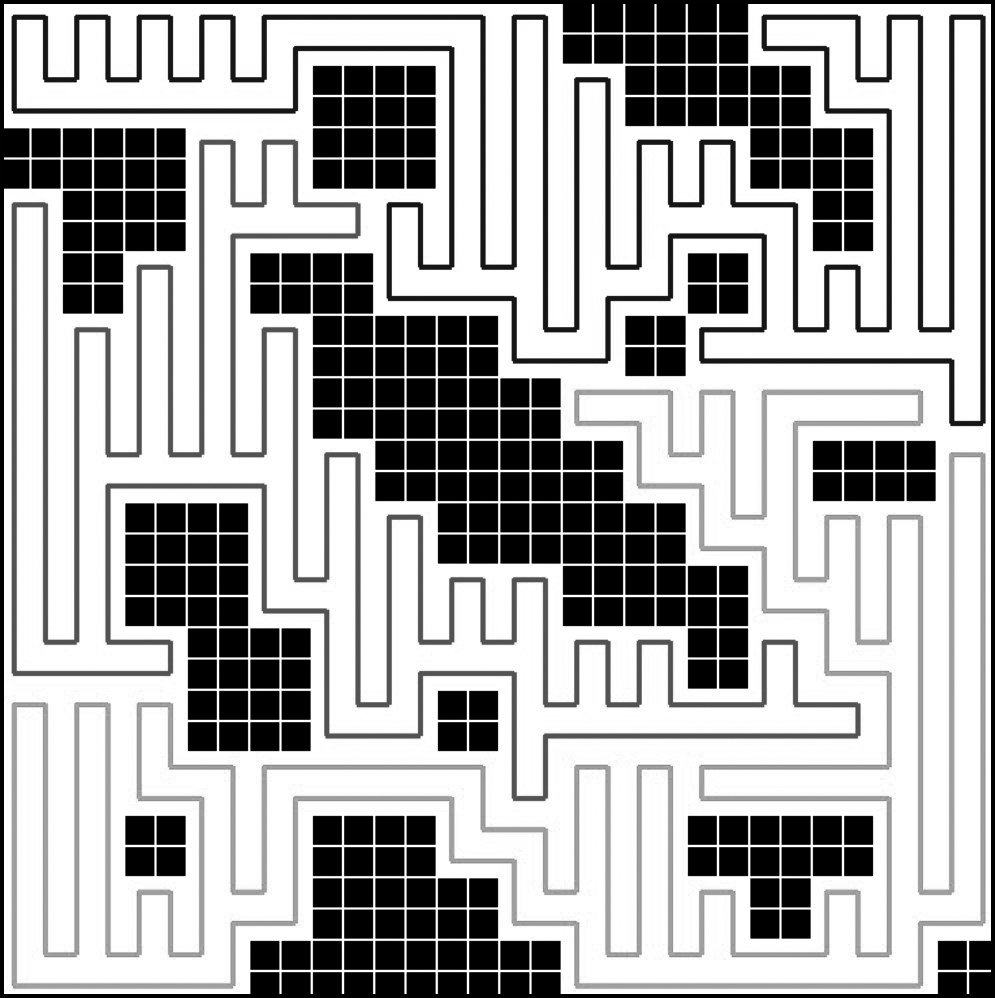}}
     \subfigure[UF-STC algorithm simulation result 2]{
    \label{Fig7.sub.4}
    \includegraphics[width=5cm,height=5cm]{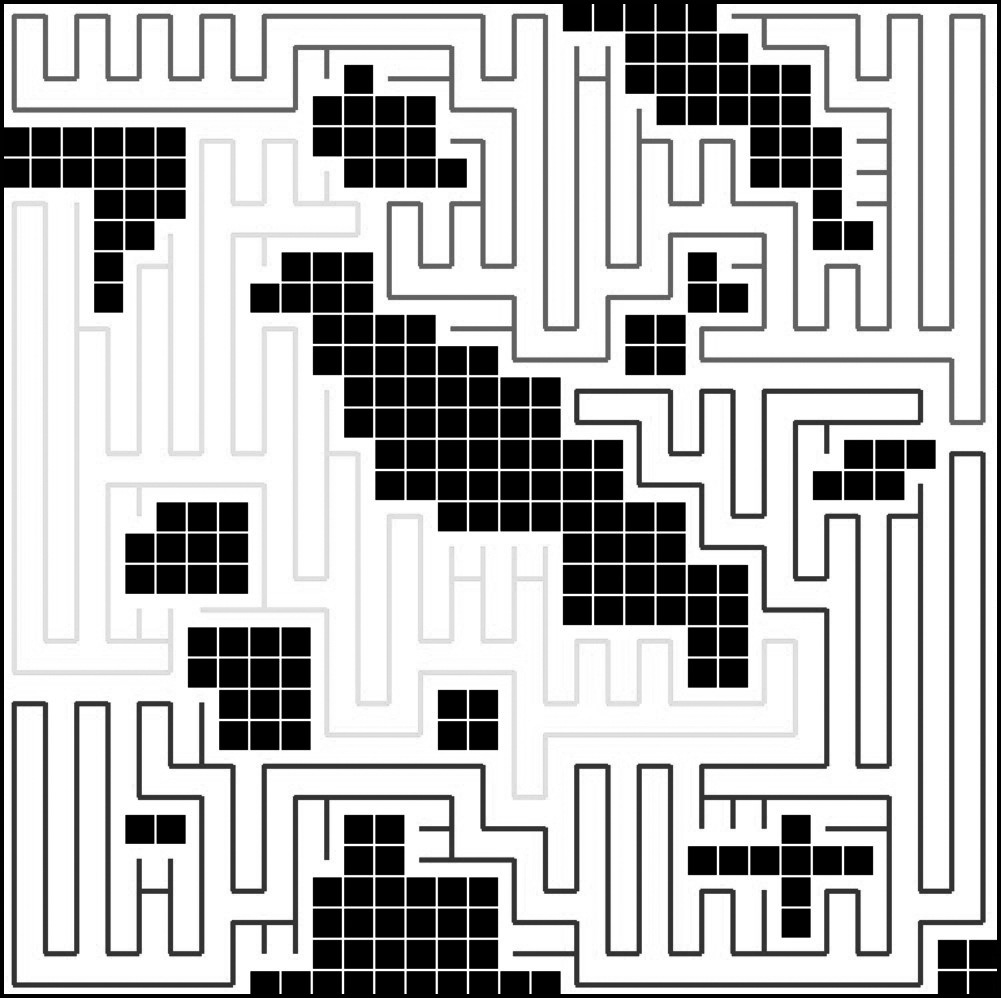}}
    \caption{STC algorithm/UF-STC algorithm simulation results}
    \label{Fig7}
\end{figure}
As shown in Figure 7, Figure 7(a) and 7(c) are the performances of traditional STC algorithm, whose path can cover all cells except small cells, which have been marked as obstacles. Figures 7(b) and 7(d) are the performances of UF-STC algorithm. It can be seen from the comparison with Figure 7(a) and 7(b) that this algorithm marks small cells as feasible are-as and covers them. Specifically, when the path covers cells adjacent to small cells, it branches out in the direction of small cells and covers them according to the method formulated by UF algorithm. STC algorithm makes the path of a single robot a closed curve, while UF algorithm makes “spikes” grow on this closed curve so as to cover small cells.

\begin{table}[htb]
\tbl{ coverage rate comparison}
{
\begin{tabular}{cccc}
\hline
Map & Robot & \begin{tabular}[c]{@{}l@{}}STC\\ Coverage   rate\end{tabular} & \begin{tabular}[c]{@{}l@{}}UF-STC\\ Coverage   rate\end{tabular} \\ \hline
d   & 3     & 97.6\%                                                               & 100\%                                                                   \\
e   & 3     & 93.0\%                                                               & 100\%                                                                   \\ \hline
\end{tabular}}
\end{table}

Table 4 shows a comparison of coverage rate between STC and UF-STC. It can be seen that when small cells exist, STC cannot achieve full coverage, while UF-STC can achieve 100\% coverage. This means that coverage can be completed with precision at robot size level which greatly improves coverage efficiency and accuracy.

\section{Conclusion and Future Work}
This paper proposes a multi-robot full coverage path planning algorithm based on improved DARP algorithm. The map in-formation expression method in this paper is grid method, which converts real maps into grid maps for analysis. This paper’s algorithm uses A* algorithm to optimize the evaluation matrix of DARP algorithm, divides the task area of each robot, and then uses UF algorithm to make up for the defect that STC algorithm cannot achieve 100\% coverage rate, and performs path planning and coverage.

Comparative analysis between A*-DARP algorithm and DARP algorithm shows that A*-DARP algorithm can effectively reduce the number of turns of robots in the path, short-en the time required for coverage, and make task allocation between robot clusters more even. This improves coverage efficiency and reduces energy loss and mechanical loss of robots.

Comparing UF-STC algorithm with STC algorithm shows that UF-STC algorithm can complete coverage at robot size level scale, effectively making up for the defect that STC algorithm cannot cover small cells, increasing coverage rate to 100\%, and greatly improving coverage accuracy.

However, there are still some problems with the proposed algorithm in this paper: 
\begin{romanlist}[(ii)]
\item For extreme map conditions or situations where robot clustering is too high, there may be situations where full coverage tasks cannot be achieved theoretically. This can be improved by adjusting the relative positions of robots;
\item For situations where environmental information is unknown, this algorithm cannot perform reasonable task allocation and path planning. These issues still need further research and discussion.
\end{romanlist}


\begin{thebibliography}{0}
\bibitem{1} Zixing C, Yian C., Survey of Multi-Robot Coverage, {\small\it
Control and Decision} {\bf 23}:481--486, 2008.
\bibitem{2} Elfes A, Moravec H., High resolution maps from wide angle sonar, {\small\it
IEEE Conference In Robotics and Automa-tion} 116--121, 1985.
\bibitem{3}	Wong S C, MacDonald B A.,A topological coverage algorithm for mobile robots , {\small\it
In Proceedings 2003 IEEE/RSJ International Conference on Intelligent Robots and Systems (IROS 2003)(Cat. No. 03CH37453)}1685--1690, 2003.
\bibitem{4} Acar E U, Choset H, Rizzi A A, et al., Morse decompositions for coverage tasks, {\small\it
The International Journal of Robotics Research} {\bf 21 (4)}: 331--344, 2002.
\bibitem{5} Guruprasad K, Dasgupta P., Distributed Voronoi partitioning for multirobot systems with limited range sensors , {\small\it
In 2012 IEEE/RSJ international conference on intelligent robots and systems}3546--3552, 2012.
\bibitem{6} Gabriely Y, Rimon E., Spanning-tree based coverage of continuous areas by a mobile robot, {\small\it
Annals of Mathematics and Artificial Intelli-gence} {\bf 31 (1-4)}:77--98, 2001.
\bibitem{7} Hart P E, Nilsson N J, Raphael B., A formal basis for the heuristic determination of minimum cost paths, {\small\it
IEEE Transactions On Systems Science and Cybernetics} {\bf 4 (2)}:100--107, 1968.
\bibitem{8} Goldberg D E, Holland J H.,Genetic algorithms and machine learning , {\small\it
Machine Learning} {\bf 3 (2)}:95--99, 1988.
\bibitem{9} Zhang Y, Liu S-H.,urvey of multi-robot task allocation, {\small\it
CAAI Trans-1 actions on Intelligent Systems} {\bf 3 (2)}:115--120, 2008.
\bibitem{10} Gerkey B P, Mataric M J., Auction methods for multirobot coordination, {\small\it
IEEE transactions on robotics and automation} {\bf 18 (5)}:758--768, 2002.
\bibitem{11} Ding Y, Yingying and He, Jiang J. ,Multi-robot cooperation method based on the ant algorithm , {\small\it
In Proceedings of the 2003 IEEE Swarm Intel-ligence Symposium. SIS'03 (Cat. No. 03EX706)} 14--18, 2003.
\bibitem{12} Athanasios Ch. Kapoutsis · Savvas A. Chat-zichristofis · Elias B. Kosmatopoulos.,DARP: Divide Areas Algorithm for Optimal Multi-Robot Coverage Path Planning, {\small\it
 Intell Robot Syst} {\bf 86}:663--680, 2017.
\bibitem{13} Nedjati A, Izbirak G, Vizvari B, Arkat J,Complete coverage path planning for a mul-ti-UAV response system in post-earthquake assessment, {\small\it
 Robotics} {\bf 5 (4)}:26, 2016.
\bibitem{14} Wang Xiao-hong, Ye Tao.,Research on robot path planning based on improved A* algorithm, {\small\it
Computer Measurement \& Control} {\bf 26 (7)}:282--286, 2018.
\bibitem{15} Wang Miao-chi, Path planning of mobile robot based on A* algorithm, {\small\it
Shenyang University of Technology} 2017.
\bibitem{16} Shi Hui, Cao Wen, Zhu Shu-lang, Application of an improved A* algorithm in shortest route planning, {\small\it
Geomatics \& Spatial Information Technology} {\bf 32(6)}:208--211, 2009.

\bibitem{17} Wang Hong wei, Ma Yong, Xie Yong, Mobile robot optimal path planning based on smoothing A* algorithm, {\small\it
Journal of Tongji University (Natural Science Edition)} {\bf 38(11)}:1647--1650, 2010.

\bibitem{18} Howie Choset, Coverage for robotics-A survey of recent results, {\small\it
Annals of Mathematics and Artificial Intelli-gence} {\bf 31(1-4)}:113--126, 2001.
\bibitem{19} Dijkstra, E., A note on two problems in connection with graphs, {\small\it
Numerische Mathematik} {\bf 1}:269--271, 1959.
\bibitem{20} Tarjan, R.E., Data Structures and Network Algorithms, {\small\it
 vol. 14. SIAM} 1983.
\bibitem{21} Wright, S.J., Coordinate descent algorithms, {\small\it
Math. Program} {\bf 151(1)}:3--34, 2015.
\bibitem{22} Hazon, N., Kaminka, G., et al, Redundancy, efficiency and robustness in multi-robot coverage, {\small\it
Proceedings of the 2005 IEEE International Conference on Robotics and Automation}, Barcelona, pp.~735–-741, 2005.
\end{thebibliography}
\end{document}